\documentclass[journal]{IEEEtran}
\newcommand{\ignore}[1]{}

\usepackage[utf8]{inputenc}
\usepackage[T1]{fontenc}
\usepackage{amsmath,epsfig}
\usepackage{url}
\usepackage{graphicx}
\usepackage{bm}
\usepackage{amssymb}
\usepackage{cite}
\usepackage{array}
\usepackage{color}
\usepackage{booktabs}
\usepackage{multirow}

\usepackage{times}
\usepackage{epsfig}
\usepackage{graphicx,psfrag,epsfig}
\usepackage{amsmath}
\usepackage{amssymb}

\usepackage{multirow}
\usepackage{array}
\usepackage{caption}
\usepackage{makecell}
\usepackage{anyfontsize}
\usepackage{marvosym}

\begin{document}
	
	\title{Referring Remote Sensing Image Segmentation with Cross-view Semantics Interaction Network}
	
	\author{Jiaxing Yang, Lihe Zhang, Huchuan Lu\\
		\IEEEcompsocitemizethanks{
			\IEEEcompsocthanksitem This work was supported by the National Natural Science Foundation of China  \#62276046 and \#62431004, and the Liaoning Natural Science Foundation  \#2021-KF-12-10. Authors J. Yang, L. Zhang, and H. Lu are
			with School of Information and Communication Engineering, Dalian
			University of Technology, Dalian, China (e-mail:  jx.yang@mail.dlut.edu.cn, zhanglihe@dlut.edu.cn lhchuan@dlut.edu.cn). 
		}.
	}
	\markboth{}{}	
	\maketitle
	\begin{abstract}
		Recently, Referring Remote Sensing Image Segmentation (RRSIS) has aroused wide attention. To handle drastic scale variation of remote targets, existing methods only use the full image as input and nest the saliency-preferring techniques of cross-scale information interaction into traditional single-view structure. Although effective for visually salient targets, they still struggle in handling tiny, ambiguous ones in lots of real scenarios. In this work, we instead propose a paralleled yet unified segmentation framework Cross-view Semantics Interaction Network (CSINet) to solve the limitations. Motivated by human behavior in observing targets of interest, the network orchestrates visual cues from remote and close distances to conduct synergistic prediction. In its every encoding stage, a Cross-View Window-attention module (CVWin) is utilized to supplement global and local semantics into close-view and remote-view branch features, finally promoting the unified representation of feature in every encoding stage. In addition, we develop a Collaboratively Dilated Attention enhanced Decoder (CDAD) to mine the orientation property of target and meanwhile integrate cross-view multiscale features. The proposed network seamlessly enhances the exploitation of global and local semantics, achieving significant improvements over others while maintaining satisfactory speed.
	\end{abstract}
	
	
	\IEEEpeerreviewmaketitle
	\section{Introduction}
	\label{sec:intro}
	Referring Remote Sensing Image Segmentation (RRSIS) has emerged as a promising technique with broad applications in agricultural production, urban planning, environmental monitoring, and land use analysis. This cross-modal task requires methods precisely to segment described targets in aerial imagery based on the given linguistic expression. As a relatively new task, it currently faces three fundamental challenges: (1) The targets exhibit drastic size variation across different remote imagery, from visually large to tiny; (2) The targets are usually surrounded by background with similar outlook. The network has to more efficiently exploit global and local contexts; (3) The targets often have long and slim shapes, which tend to distribute along different directions, for example, road, ship, and overpass. Their orientation-diverse semantics should be more mined in model designing.
	\begin{figure}[tbp]
		\centering
		\includegraphics[scale=0.52]{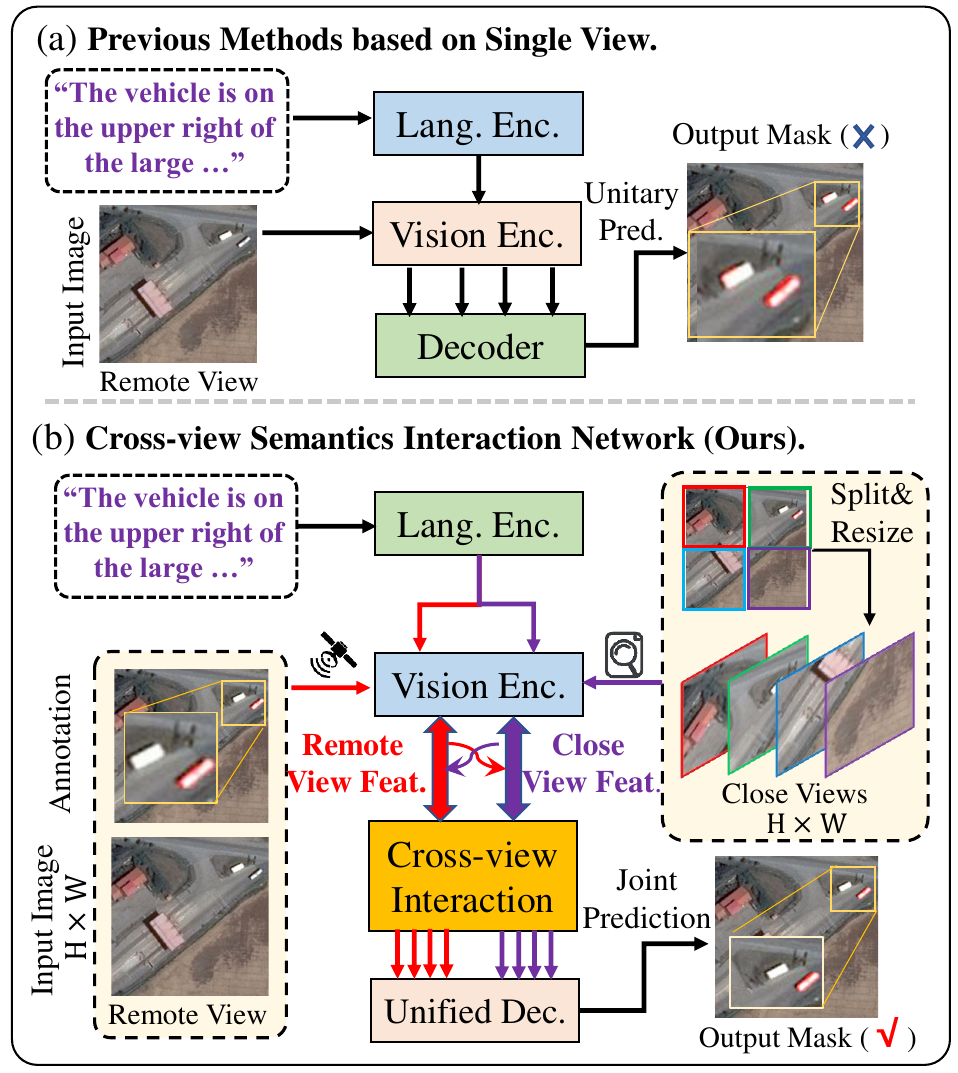}
		\caption{Architectures for referring remote sensing image segmentation. (a) Single-view methods; (b) Cross-view CSINet.}
		\label{fig:architecture_comparision}
	\end{figure}

	Inspired by the achievements in referring image segmentation for natural targets (RIS)~\cite{hu2016segmentation, li2018referring, liu2017recurrent, mdetr, shi2018key, yelinwei2019cross, hui2020linguistic, vlt, hu2020bi123, yu2018mattnet, luo2020multi, lts, Liu2022InstanceSpecificFP, liu2023polyformer,Kim2022ReSTRCR, ding2023bilateral, feng2021encoder,yang2022lavt,yang2023semantics, yang2023referring, tang2023contrastive, wu2022towards,li2023fully,shang2022cross,CMIRNet}, recent methods~\cite{lei2024exploring, yuan2024rrsis, danet} propose various customized solutions for RRSIS. Some of them, such as LGCE~\cite{yuan2024rrsis}, RMSIN~\cite{liu2024rotated}, and CroBIM~\cite{crobim}, design various multiscale feature interaction techniques to excavate the scale variation information. Others, such as FIANet~\cite{lei2024exploring} and DANet~\cite{danet}, make their efforts in promoting the semantics consistency of language and vision modalities. Although great progress has been made, they still face critical limitations. The first is that they constrain their solutions
	to traditional encoder-decoder structure of single view (See Figure \ref{fig:architecture_comparision}(a)), failing to strike a balance in depicting  global context of wide field of view and capturing small and weak details. 
	Even in their cross-scale feature interaction, early vision features are excluded (LGCENet~\cite{yuan2024rrsis}) or severely transformed (RMSIN~\cite{liu2024rotated}, FIANet~\cite{lei2024exploring}, CroBIM~\cite{crobim}) to focus the attention on extremely big target,  causing their performance worse in segmenting out tiny and ambiguous targets. The second is that existing method~\cite{liu2024rotated} heavily relies on squared CNN based techniques that however are weak in capturing the full appearance of long and slim remote target, as the static parameter can not be content-adaptive and its probing scope is constrained. Therefore, more advanced way to achieve high-quality RRSIS still needs exploring.
	
	In this work, we propose a paralleled yet unified segmentation framework Cross-view Semantics Interaction Network (CSINet), which feeds on low-resolution complete image in remote-view branch and high-resolution image patches in close-view branch to effectively handle targets of different sizes,    
	as seen in Figure \ref{fig:architecture_comparision}(b). The features from distant view own global representative ability, while those from close view tend to contain rich details. The network mimics human behavior in observing vague and tiny target of interest, and orchestrates visual cues from both remote and close distances to capture target appearances. 
	Inspired by the window module of Swin Transformer~\cite{swin}, a Cross-View Window-attention module (CVWin) is engineered to achieve information communication between different view branches to promote the unified feature representation in every encoding stage.
	In addition, a Collaboratively Dilated Attention enhanced Decoder (CDAD) is proposed to integrate cross-view multi-scale features and to simultaneously enhance the network's ability to mine orientation property of target. In the decoder, a novel collaboratively dilated attention is designed to enhance final-stage cross-modality features of both remote and close views. 
	Our main contributions can be summarized as:
	
	\begin{itemize}
		\item We design a novel segmentation framework Cross-view Semantics Interaction Network (CSINet) for RRSIS, simultaneously exploiting rich global and local cues contained in remote and close views under a patalleled yet unified paradigm. It exhibits competitive performance in handling targets with broad scale diversity and strong generalization ability on the three challenging datasets, namely RRSIS-D, RefSegRS, and RIS-Bench.
		
		\item We propose a Cross-View Window-attention module (CVWin) to achieve information communication between cross-modality features from different views, finally promoting their representational ability for remote target.
		
		\item We propose a Collaboratively Dilated Attention enhanced Decoder (CDAD) to integrate cross-view multiscale feature and simultaneously enhance the network's ability to mine the orientation property of target.
	\end{itemize}

	\section{Related Works}
	\subsection{Referring Image Segmentation.}
    Referring Image Segmentation (RIS)~\cite{hu2016segmentation, liu2017recurrent, li2018referring, qiu2019referring, hui2020linguistic, mdetr, feng2021encoder, yang2023referring, yang2022lavt,yang2023semantics,tang2023contrastive, wu2022towards, liu2023polyformer, lai2024lisa} aims to segment visual targets in natural images based on linguistic descriptions, which typically specify target attributes such as actions, categories, spatial relationships, and visual characteristics. The task coupled with video-text retrieval~\cite{chen2020fine, wen2020learning}, image captioning~\cite{xu2015show, yu2019multimodal}, and language-guided object segmentation/grounding~\cite{hu2016segmentation, li2021transformer, zhang2020language} together plays an indispensable role in human-robot interaction. Early approaches~\cite{hu2016segmentation, liu2017recurrent, li2018referring, qiu2019referring, hui2020linguistic, mdetr, feng2021encoder} predominantly rely on CNN-based techniques. As the first one posing the task, Hu \emph{et al.} in \cite{hu2016segmentation} introduced a baseline architecture where sentence-level embeddings are broadcasted to all spatial positions of visual features, generating coarse cross-modal representations for segmentation through an FPN decoder. Subsequent works focus on refining cross-modal alignment mechanisms. In KWAN~\cite{shi2018key}, Shi \emph{et al.} employed key-word-aware visual context module to model the visual context among multiple image regions in accordance with corresponding key words, and resorted to an MLP segmentation head that feeds on vision features, key-word-aware visual context features, and key-word-weight query features to predict the final mask. In LTS~\cite{lts}, Jing \emph{et al.} adopted a coarse-to-fine paradigm to achieve RIS, first generating visual priors through cross-modal localization and then refining segmentation through spatial-semantic guidance.
    In  CEFNet~\cite{feng2021encoder}, Feng \emph{et al.} exploited the inherent structure of ResNet to make the cross-modality happen alongside the vision feature encoding process, gradually forcing vision and language-aligned features to approach each other in semantics space.
    Besides, a co-attention fusion module were designed to update cross-modal features and promote their semantics consistency. However, although these works have achieved significant improvements, their heavy reliance on CNN causes poor performance when handling complex scenarios, as the global representational and cross-modality modeling abilities of static kernels are weak.
	
	To overcome, recent methods~\cite{vlt, yang2022lavt, yang2023semantics, tang2023contrastive, wu2022towards, magnet, lai2024lisa} introduce various transformer-dominating techniques into the RIS task, achieving a remarkable improvements over the previous methods. In LAVT~\cite{yang2022lavt}, Yang \emph{et al.} designed a gating mechanism and successfully integrated the language information into encoding stages of Swin Transformer~\cite{swin}. Over LAVT, SADLR~\cite{yang2023semantics} gradually introduce more-refined localization cues in its repeated decoding process. In CGFormer~\cite{tang2023contrastive}, Tang \emph{et al.} proposed a group transformer cooperated with contrastive learning to achieve object-aware cross-modal reasoning, explicitly grouping visual features into different regions and modeling their dependencies conditioning on language features. In PolyFormer~\cite{liu2023polyformer}, Liu \emph{et al.} parsed segmentation of referent prediction of sequential points, formulating geometric localization as a regression task. In LISA~\cite{lai2024lisa}, Lai \emph{et al.} employed an embedding-as-mask paradigm to incorporate
	new segmentation capabilities into the multi-modal large language model LoRA~\cite{hu2022lora}.
	Because of domain gap between aerial and natural scenes, rough transferring these methods to remote sensing images
	is problematic. Peculiar properties of referred remote target in spatial distribution should be more mined.
	
	\subsection{Referring Remote Sensing Image Comprehension.}
	Referring Remote Sensing Image Comprehension involves subtasks of referring remote sensing image detection (RRSID) and segmentation (RRSIS). The RRSID requires grounding the described object using a bounding box. In GeoVG~\cite{sun2022visual},  Sun \emph{et al.} first proposed a large-scale dataset for remote sensing target detection, and a baseline method consisting of adaptive region attention and context fusion modules. In RSVG~\cite{zhan2023rsvg}, Yang \emph{et al.} devised a transformer-based multi-granularity visual-language
	fusion module to incorporate effective scale-variation information from cross-modality features. In LPVA~\cite{lpva}, Li \emph{et al.} proposed a language-guided progressive attention module to address the attention drift problem, and a multi-level feature enhanced decoder to aggregate visual contextual information to achieve more accurate localization. 
	
	The RRSIS requires segmenting out the described region at the pixel level. Some efforts~\cite{yuan2024rrsis, liu2024rotated, danet, lei2024exploring, crobim} have been made to achieve the task so far. It is first explored in LGCE~\cite{yuan2024rrsis}, in which Yuan \emph{et al.} proposed the first dataset RefSegRS and a transformer-based baseline method, which was adapted from LAVT~\cite{yang2022lavt}. In addition to developing another dataset RRSIS-D that contains targets with more diverse scales, Liu \emph{et al.} in RMSIN~\cite{liu2024rotated} proposed an intra-scale interaction module and a cross-scale interaction module to handle fine-grained information within and across feature of different scale. Besides, an adaptive rotated convolution enhanced decoder is applied to enhance the network's understanding of object orientation. In DANet~\cite{danet}, Pan \emph{et al.} proposed a dual alignment network and an explicit affinity alignment strategy to mitigate the domain discrepancies. In FIANet~\cite{lei2024exploring}, Lei \emph{et al.} adopted a fine-grained image-text alignment to promote the cross-modality consistency. The noun and position components in the language pivotal to target are specifically considered in cross-modality relationship modeling. In CroBIM~\cite{crobim}, Dong \emph{et al.} engineered a context-aware prompt modulation module to integrate multi-scale visual context with prompts generated from linguistic features. Besides, a language feature flows up module was also proposed to facilitate
	interaction between visual and linguistic features across
	multiple scales. Another large-scale benchmark RIS-Bench was proposed in~\cite{crobim}, where language expressions describe more complex geospatial properties.
	
	However, although significant improvements have been made, current RRSIS methods~\cite{yuan2024rrsis, liu2024rotated, danet, lei2024exploring, crobim} still suffer from inherent limitations. Their exclusive reliance on remote-view input leads to inadequate feature representation for fine-grained details. Moreover, their cross-scale interaction mechanisms, based on downsampled low-resolution features, exhibit salient target bias that particularly compromises segmentation accuracy for non-salient objects. To overcome these challenges, we propose a synergistic dual-stream architecture that strategically combines global semantic-strong features from the remote view and local semantic-strong features from the close-view observation, establishing a stage-by-stage mutual reinforcement mechanism to capture object's appearance with distinct scales in a unified framework.
	\begin{figure*}[htbp]
		\centering
		\includegraphics[scale=0.63]{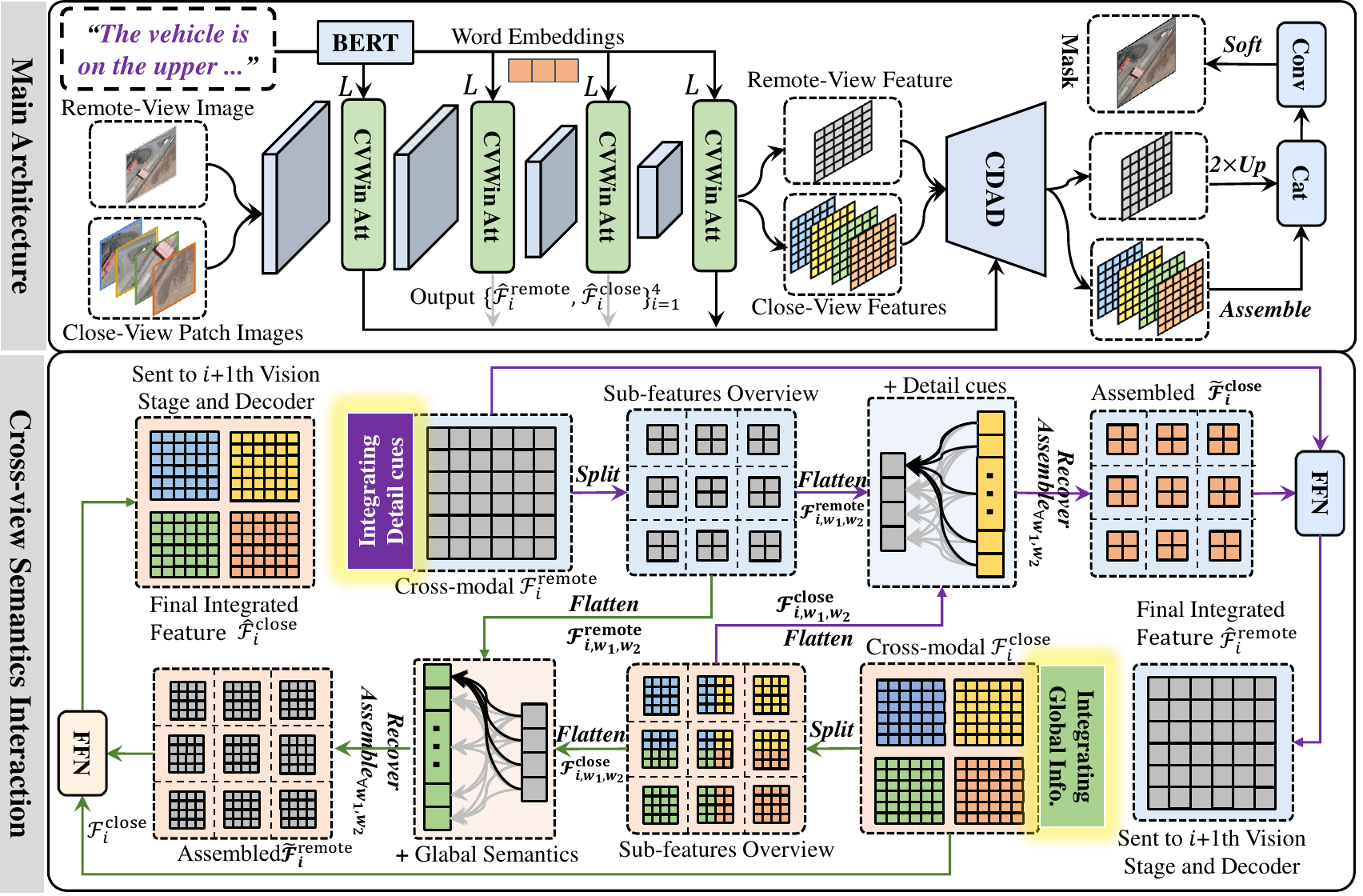}
		\caption{Architecture overview of CSINet. It feeds on a complete image that simulates remote view and several image patches that simulate close view to coordinate global and local information to predict the target's appearance for RRSIS, where CVWin is employed in every vision encoding stage to promote information communication and CDAD is used to decode. In the bottom, cross-view information interaction of CVWin's second step is sketched, corresponding to Eq. \eqref{from-close--remote}-\eqref{method:remote2close}.} 
		\label{Architecture}
	\end{figure*}
	
	\section{Methodology}
	\label{sec:methods}
	The architecture of cross-view semantics interaction network (CSINet) is introduced in Sec.~\ref{sec:overall_architecture}. The details of cross-view window-attention module (CVWin) are introduced in Sec. \ref{sec:cwwin}. Finally, collaboratively dilated attention enhanced decoder (CDAD) are introduced in Sec. \ref{sec:cdad}. 
	

	\subsection{Overall Architecture.}
	\label{sec:overall_architecture}
	The overall structure is visualized in Figure \ref{Architecture}.  Following  LGCE~\cite{yuan2024rrsis} and RMSIN~\cite{liu2024rotated}, Swin Transformer~\cite{swin} and BERT~\cite{bert} are utilized to extract vision and language features, respectively. 
	For vision inputs, the network feeds on low-resolution image $\cal{I}^{\rm{remote}}$ that is resized to size $\rm{H}\times \rm{W} \times 3$ from the original image to simulate the \textbf{REMOTE} view, and a set of non-overlapping image patches $\{\mathcal{I}_{n1,n2}^{\rm{close}}\}_{n1,n2=1}^{\rm{N^{view}}}$ split from high-resolution image of size $\rm{N^{view}}\rm{H}\times \rm{N^{view}}\rm{W} \times 3$ that is also resized from the original image to simulate the \textbf{CLOSE} view. $\rm H$ and $\rm W$ denote the height and width of image, respectively, and $\rm{N^{view}}$  specifies the number of image patches. Each image patch keeps the same spatial resolution as that of remote-view image. Empirically, the super-parameter $\rm{N}^{\rm{view}}$ is set to 2. The remote-view image and close-view patches are concatenated and fed into vision encoder. In the $i$-th encoding stage ($i=1,...,4$), vision features extracted from the remote-view image and the $(n1, n2)$-th close-view patch are denoted as $\mathcal{V}^{\rm{remote}}_{i}$ and $\mathcal{V}_{i, n1, n2}^{\rm{close}}$. Their size is $\rm{H}_{\it{i}} \times \rm{W}_{\it{i}} \times \rm{C}_{\it{i}}^{vis}$, with $\rm{H}_{\it{i}} = H/2^{(\it{i}+\rm{1})}$ and $\rm{W}_{\it{i}}= W/2^{(\it{i}+\rm{1})}$. Language feature $T$ has size of $\rm L\times \rm C^{lang}$, where $\rm L$ specifies the  length of words.
	
	The proposed CVWin is installed in-between vision encoding stages to supplement not only the remote-view feature with detail cues, but also the close-view feature with global semantics. The stage-by-stage employment of CVWin will progressively force the extracted features to more focus their attention on the position and detail of right candidate. The enhanced features by CVWin are integrated into the next vision feature encoding stage and simultaneously sent to CDAD decoder.
	The final mask is generated using a simple convolution over the concatenation of final remote-view feature and spliced close-view feature.
	
	\subsection{Cross-View Window Attention Module.}
	\label{sec:cwwin}
	In CVWin, we first generate cross-modality features for image and each patch, and then conduct cross-view semantics interaction, in which an asymmetrical-window cross-attention method is developed (as seen in the bottom of Figure \ref{Architecture}). The attention keeps the number of windows consistent across remote view and spliced close-view features, initiating information communication between every pair of corresponding windows. Compared to vanilla seq2seq cross-attention method and its based downsampling variants, it makes the computational overhead controlled while protecting the contextual information of each-view feature from severely damaging. The tokens in close-view window, and their counterparts in remote-view window overall refer to the same image content, objectively modeling local token's dependencies alongside the cross-view information communication process. After split, the enhanced close-view features coupled with that of remote-view together can undergo further paralleled encoding.
	
	\subsubsection{Cross-modality Feature Generation}
	The remote-view vision feature in the $i$th stage and language feature will be first processed as follows:
	\begin{equation} \label{eq:cross_modal_remote_qkv}
		\begin{aligned}
			& Q^{\rm{remote}} = Flatten_{12}(Conv2D(\mathcal{V}^{\rm{remote}}_{i})),\hspace{3pt} \\
			& K^{\rm{lang}} = Perm_{21}(Linear(T)),\hspace{3pt} V^{\rm{lang}} =Linear(T),\hspace{3pt}
		\end{aligned} 
	\end{equation}
	in which $Q^{\rm{remote}}$, $K^{\rm{lang}}$, and $V^{\rm{lang}}$ represent query, key, and value entities, respectively.  $Conv2D$ represents $1\times 1$ convolution with stride of 1 to recombine channel information.  $Flatten_{12}$ represents flattening feature in a first-row-then-column manner.  $Linear$ adjusts the channel dimension.  $Perm_{21}$ represents exchanging the first and second indices of language feature. As such, the aligned language feature can be computed as follows:
	\begin{equation} \label{eq:cross_modal_alignment}
		\small
		\begin{aligned}
			&\mathcal{F}_i^{\rm{remote}} = Recover(Soft_2(\frac{Q^{\rm{remote}}@ K^{\rm{lang}}}{\sqrt{\rm{C^{vis}_{\textit{i}}}}})@V^{\rm{lang}}), 
		\end{aligned} 
	\end{equation}
	in which the symbol $@$ denotes matrix multiplication.  $Soft_i$  conducts exponential normalization over the $i$th dimension.  $Recover$  performs the reverse transformation of $Flatten_{12}$.
	
	Inspired by LAVT~\cite{yang2022lavt}, we integrate the aligned language feature and vision feature using a gate mechanism as:
	\begin{equation} \label{cross-modality-gate}
		\begin{aligned}
			&\mathcal{G}_i^{\rm{remote}} = Tanh(FFN(Cat([\mathcal{F}_i^{\rm{remote}},\mathcal{V}_i^{\rm{remote}}]))), \\
			&\mathcal{F}_i^{\rm{remote}} = \mathcal{G}_i^{\rm{remote}}*FFN(Cat([\mathcal{F}_i^{\rm{remote}}, \mathcal{V}_i^{\rm{remote}}]))), \\
			&\mathcal{F}_i^{\rm{remote}} = \mathcal{V}_i^{\rm{remote}}+ FFN(\mathcal{F}_i^{\rm{remote}}),
		\end{aligned} 
	\end{equation}
	in which the manipulation  $Cat$ denotes feature concatenation along channel dimension.   $FFN$ denotes feature re-projecting  by a convolution of kernel size $1\time1$ and stride size 1  for channel dimension alignment.
	
	Vision features from close-view patches are first re-arranged into the complete one, as follows:
	\begin{equation} \label{att1}
		\begin{aligned}
			&\mathcal{V}_i^{\rm{close}} = Assemble_{\forall n1, n2}(\mathcal{V}_{i, n1, n2}^{\rm{close}}),
		\end{aligned} 
	\end{equation}
	in which the operation $Assemble_{\forall n1, n2}$ represents assembling the sub-feature to the before-splitting position of its corresponding image patch. The $\mathcal{V}_i^{\rm{close}}$ is used to query language information by replacing $\mathcal{V}_i^{\rm{remote}}$ in Eq. (\ref{eq:cross_modal_remote_qkv}) and (\ref{eq:cross_modal_alignment}). The final cross-modality feature $\mathcal{F}_i^{\rm{close}}$ is generated via Eq. (\ref{cross-modality-gate}), by replacing $\mathcal{F}_i^{\rm{remote}}$ and $\mathcal{V}_i^{\rm{remote}}$ with $\mathcal{F}_i^{\rm{close}}$ and $\mathcal{V}_i^{\rm{close}}$. 
	
	\subsubsection{Cross-view Semantics Interaction}
	Let $\rm{S}^{\rm{win}}$ denote the predefined window size hyperparameter.  To accommodate the following operations, We first utilize bilinear interpolation manipulation to upsample $\mathcal{F}_i^{\rm{remote}}$ to $\lceil {\rm{H}_{\textit{i}}}/{\rm{S}^{\rm{win}}}\rceil*\rm{S}^{\rm{win}}$ and $\mathcal{F}_i^{\rm{close}}$ to $\lceil \rm{H}_{\textit{i}}/\rm{S}^{\rm{win}}\rceil*(\rm{N}^{\rm{view}}\rm{S}^{\rm{win}})$, in which the symbol of $\lceil\cdot\rceil$ represents rounding up the float number to the recent integer. As such, the feature $\mathcal{F}_i^{\rm{remote}}$ and $\mathcal{F}_i^{\rm{close}}$ can be split into $\rm{N}^{\rm{win}}\times \rm{N}^{\rm{win}}$ sub-features, where $\rm{N}^{\rm{win}}=\lceil \rm{H}_{\textit{i}}/\rm{S}^{\rm{win}}\rceil$. Their $(w1,w2)$-th sub-features $\mathcal{F}_{i,w1,w2}^{\rm{remote}}$ and $\mathcal{F}_{i,w1,w2}^{\rm{close}}$ serve as bridge to communicate information between two branches in a bidirectional manner.
	
	The detail information is transmitted from tokens within the close-view $(w1,w2)$-th sub-feature to tokens within the remote-view $(w1,w2)$-th sub-feature. The two features are first processed as:
	\begin{align} \label{from-close--remote}
		&Q^{\rm{remote}},V^{\rm{close}} = Flatten_{12}(\mathcal{F}_{i,w1,w2}^{\rm{remote}}), Flatten_{12}(\mathcal{F}_{i,w1,w2}^{\rm{close}}) \nonumber \\ 
		&K^{\rm{close}} = Perm_{21}(Flatten_{12}(\mathcal{F}_{i,w1,w2}^{\rm{close}})), 
	\end{align}
	in which the involved manipulations can be referred to those in Eq. (1) and Eq. (2). The aligned close-view sub-feature can be  computed as follows:
	\begin{equation} \label{method:close2remote}
    \small
		\begin{aligned}
			&\mathcal{\tilde{F}}_{i,w1,w2}^{\rm{close}} = Recover(Soft_2(\frac{Q^{\rm{remote}}@ K^{\rm{close}}}{\sqrt{\rm{C^{vis}_{\it{i}}}}})@V^{\rm{close}}), 
		\end{aligned} 
	\end{equation}
	in which $w1,w2=1,..,\rm{N}^{\rm{win}}$. The sub-features then can be grouped back into their before-window-splitting positions again as the complete one as follows:
	\begin{equation} \label{method:close2remote_assemble}
		\begin{aligned}
			\mathcal{\tilde{F}}_{\textit{i}}^{\rm{close}} = Assemble_{\forall{w1,w2}}(\mathcal{\tilde{F}}_{i,w1,w2}^{\rm{close}}).
		\end{aligned} 
	\end{equation}
	
    The final feature $\mathcal{\hat{F}}_i^{\rm{remote}}$ for the remote-view branch is derived by integrating $\mathcal{F}_{i}^{\rm{remote}}$ and $\mathcal{\tilde{F}}_{i}^{\rm{close}}$ via $FFN$. Global semantics from tokens within the $(w1,w2)$-th remote-view sub-feature is transmitted to tokens within the close-view $(w1,w2)$-th sub-feature. Primarily, the following matrices are computed as:
	\begin{align}
		&Q^{\rm{close}}, V^{\rm{remote}}= Flatten_{12}(\mathcal{F}_{i,w1,w2}^{\rm{close}}),Flatten_{12}(\mathcal{F}_{i,w1,w2}^{\rm{remote}}) \nonumber\\
		&K^{\rm{remote}} = Perm_{21}(Flatten_{12}(\mathcal{F}_{i,w1,w2}^{\rm{remote}})).\hspace{3pt} 
	\end{align} 
	
    The aligned remote-view feature can be denoted as:
	\begin{align} \label{method:remote2close}
		&\mathcal{\tilde{F}}_{i, w1, w2}^{\rm{remote}} = Recover(Soft_2(\frac{Q^{\rm{close}}@ K^{\rm{remote}}}{\sqrt{\rm{C^{vis}_{\it{i}}}}})@V^{\rm{remote}}), \nonumber\\ 
		&\mathcal{\tilde{F}}_{i}^{\rm{remote}} = Assemble_{\forall{w1,w2}}(\mathcal{\tilde{F}}_{i,w1,w2}^{\rm{remote}}).
	\end{align}
	
    Then the final feature $\mathcal{\hat{F}}_i^{\rm{close}}$ for the close-view branch can be derived by integrating $\mathcal{F}_{i}^{\rm{close}}$ and $\mathcal{\tilde{F}}_{i}^{\rm{remote}}$ via $FFN$. After CVWin's cross-view interaction, the feature $\mathcal{\hat{F}}_i^{\rm{close}}$ is again split into $(\rm{N}^{view}, \rm{N}^{view})$ patches to keep the same spatial resolution as $\mathcal{\hat{F}}_i^{\rm{remote}}$ for further parallel processing.

	\subsection{Collaboratively Dilated Attention Enhanced Decoder.}
	\label{sec:cdad} 
\begin{figure}[htbp]
		\centering
		\includegraphics[scale=0.55]{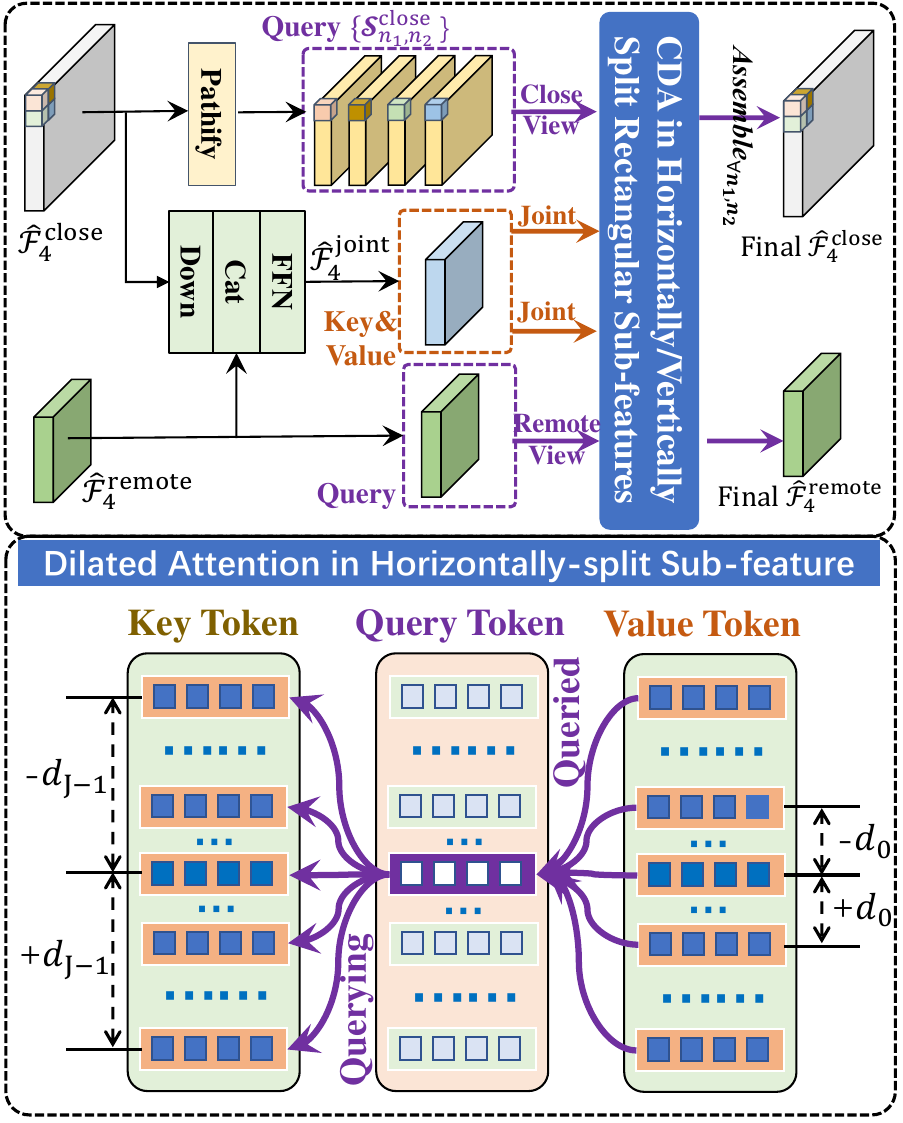}
		\caption{The details of CDA. The upper part shows how remote-view and close-view features retrieve information from their fused feature $\mathcal{\hat{F}}_4^{\rm{joint}}$. The bottom part shows how tokens in each row within the horizontally-split rectangular sub-feature aggregate information from its counterpart in $\mathcal{\hat{F}}_4^{\rm{joint}}$ dilatedly. } 
		\label{fig:CDA}
	\end{figure}

    Conventional stacked CNN architectures face three inherent limitations when processing slender, oriented targets: 1) rigid spatial constraints from fixed kernel configurations; 2) insufficient target-aware reasoning at the object level; 3) progressive information degradation through standard decoding operations. To address these limitations, we employ Collaboratively Dilated Attention decoder (CDAD), in which a content-adaptive Collaboratively Dilated Attention (CDA, see Figure \ref{fig:CDA}) mechanism that enables the mining of orientation-diverse target semantics in a unified manner. The final prediction is generated through synergistic integration of multi-view features from both remote and close observation ranges, namely $\{\mathcal{\tilde{F}}_{i}^{\rm{remote}}\}_{i=1}^{4}$ and $\{\mathcal{\tilde{F}}_{i}^{\rm{close}}\}_{i=1}^{4}$. To reduce computational overhead and exploit the strong representational ability of top-level cross-modality features, the attention leverages $\mathcal{\tilde{F}}_{4}^{\rm{remote}}$ and $\mathcal{\tilde{F}}_{4}^{\rm{close}}$ to establish dynamic associations between feature elements based on determined dilation distances that match typical target distributions. For each horizontally-split sub-feature ($\rm{H}_{4}\times \rm{S}^{\rm{slice}}$), row-wise token interactions are first computed through dilated column queries. Spatial relationships are then completed through dimension-transposed processing, achieved by exchanging row/column indices of previous product and repeating the similar attention protocol. The specific processes are omitted for convenience. The decoder systematically combines refined cross-view features through residual connections while preserving scale-specific characteristics.
    
	
	\textbf{Collaboratively Dilated Attention:}
	To be specific, $\mathcal{\hat{F}}_4^{\rm{remote}}$ is first upsampled via bilinear interpolation to $\rm{H}_{4}^{adjust}$ along both height and width dimensions,
	where $\rm{H}_{4}^{adjust} =\rm{N}^{\rm{slice}}\rm{S}^{\rm{slice}}$ with $\rm{N}^{\rm{slice}}=\lceil {\rm{H}_{4}}/{\rm{S}^{\rm{slice}}}\rceil$.
    Then the multi-view joint feature $\mathcal{\hat{F}}_4^{\rm{joint}}$ is calculated by a series of manipulations of downsampling $Down(\mathcal{\hat{F}}_4^{\rm{close}},\rm{H}_{4}^{adjust})$, concatenation $Cat$ and channel dimension reduction $FFN$.
    As such, $\mathcal{\hat{F}}_4^{\rm{remote}}$ from remote view can query information from $\mathcal{\hat{F}}_4^{\rm{joint}}$. The matrices $\mathcal{Q}$, $\mathcal{K}$ and $\mathcal{V}$ are computed:
	\begin{equation} \label{query_key_value}
		\begin{aligned}
			& \mathcal{Q}^{\rm{remote}} = Split(ConvQ_{1\times 1}(\mathcal{\hat{F}}_4^{\rm{remote}}+\mathcal{P})),\\
			& \mathcal{K}^{\rm{joint}} = Split(ConvK_{1\times 1}(\mathcal{\hat{F}}_4^{\rm{joint}}+\mathcal{P})), \\
			&\mathcal{V}^{\rm{joint}} = Split(ConvV_{1\times 1}(\mathcal{\hat{F}}_4^{\rm{joint}})),
		\end{aligned} 
	\end{equation}
	in which  $ConvQ_{1\times 1}$ represents $1\times 1$ convolution of stride 1 and padding size $(0,0)$.  $ConvK_{1\times 1}$ and $ConvV_{1\times 1}$  represents the $1\times 1$ convolution  of stride 1 and padding size $(\rm{H}_{4}^{adjust},0)$. The positional tensor $\mathcal{P}$ is generated following previous works~\cite{hu2016segmentation, hu2020bi123, feng2021encoder}.   $Split$ represents splitting the features along their horizontal direction into $\rm{N}^{\rm{slice}}$ sub-features. 
    Key entity $\mathcal{K}^{joint}$ is further processes as follows:
	\begin{align}\label{slice:key:gen}
		&\mathcal{K}^{\rm{joint}} = Cat(\mathcal{K}^{\rm{joint}}[\rm{H}_{4}^{adjust}: 2\rm{H}_{4}^{adjust}, :, :, :],\nonumber \\
		&\quad \mathcal{K}^{\rm{joint}}[\rm{H}_{4}^{adjust}\pm \textcolor{black}{\it{d}_{\rm 0}}: 2\rm{H}_{4}^{adjust}\pm \textcolor{black}{\it{d}_{\rm 0}}, :, :, :],...,  \\
		&\quad \mathcal{K}^{\rm{joint}}[\rm{H}_{4}^{adjust}\pm \textcolor{black}{\it{d}_{\rm J-1}}: 2\rm{H}_{4}^{adjust}\pm \textcolor{black}{\it{d}_{\rm J-1}}, :, :, :]), \nonumber
	\end{align}
	in which $d_{j} = \lfloor \rm{H}_{4}^{adjust}/{2^{(\rm{J}- \it{j})}} \rfloor \rm{,} \hspace{3pt} \it{j} = \rm{0,1,..., \rm{J-1}}$. The hyper-parameter $\rm{J}$ controls the density of queried rows in the rectangular box, and $\lfloor \cdot \rfloor$ represents rounding the number down to its closest integer.  $Cat$ represents concatenating the set of tensors along the width dimension. The generated $\mathcal{K}^{joint}$ is of size $\rm{H}_{4}^{adjust}\times \rm{N}^{\rm{slice}} \times (2\rm{J}+1)\rm{S}^{\rm{slice}} \times C_4^{\rm{vis}}$.
	
	Note that $\mathcal{V}^{joint}$ is also processed using Eq. \eqref{slice:key:gen}. The enhanced feature can be computed in the following manner:
	\begin{equation} \label{eq:row_enhancement}
		\small
		\begin{aligned}
			&\mathcal{R}_4^{\rm{remote}} = Soft_4(\frac{\mathcal{Q}^{\rm{remote}}@ Perm_{1243}(\mathcal{K}^{\rm{joint}})}{\sqrt{\rm{\rm{C}_4^{vis}}}})@ \mathcal{V}^{\rm{joint}},  \\
			&\hat{\mathcal{F}}_4^{\rm{remote}} =  \hat{\mathcal{F}}_4^{\rm{remote}}+LN(FFN(Recover(\mathcal{R}_4^{\rm{remote}}))),
		\end{aligned}
	\end{equation}
	in which  $Perm_{1243}$ represents to exchange the third and fourth indices of tensor.  $Soft_{4}$ represents to perform exponential normalization along the fourth dimension.  $\mathcal{R}_4^{\rm{remote}}$ is recovered to keep aligned to $\hat{\mathcal{F}}_4^{\rm{remote}}$ spatially. 
	
	Different form the query generation process in remote-view, the query entity for the close-view branch is generated by patchifying $\mathcal{\hat{F}}_i^{\rm{close}}$ that is upsampled to $\rm{N}^{view}\rm{H}_{4}^{adjust}$ via bilinear interpolation manipulation into a set of sub-features $\mathcal{\hat{F}}_i^{\rm{close}}[n1::\rm{N^{view}},\it{n}\rm{2}::\rm{N^{view}},:]$, in which every of the sub-feature $\mathcal{Q}_{n1,n2}^{\rm{close}}$ has the same spatial resolution to that of $\mathcal{\hat{F}}_i^{\rm{joint}}$. Each of $\{\mathcal{Q}_{n1,n2}^{\rm{close}}\}_{n1,n2=1}^{\rm{N^\text{view}}}$ coupled with $\mathcal{\hat{F}}_i^{\rm{joint}}$ can be used to achieve semantics enhancement via the pipeline defined in Eq. \eqref{query_key_value}-\eqref{eq:row_enhancement}. The products are finally grouped into their pre-patchified positions for further decoding.
	
	\textbf{Cross-view Multiscale Feature Integration:}
	  The $\mathcal{\hat{F}}_4^{\rm{close}}$ is split as the generation process of close-view image patches. The enhanced $\mathcal{\hat{F}}_4^{\rm{remote}}$ and $\{\mathcal{\hat{F}}_{4, n1, n2}^{\rm{close}}\}^{\rm{N^{\text{view}}}}_{n1, n2=1}$ are concatenated along the batch dimension, denoted as $\mathcal{D}_1$. The prediction result can be written as:
	\begin{equation} \label{remaining_decoder}
		\begin{aligned}
            &\mathcal{D}_i = CBR(Cat(\mathcal{I}_i,Up(\mathcal{D}_{1})),i=2,...,4,\\
			&\mathcal{I}_i = CBR(Cat(Up(\mathcal{D}_{i-1}),\mathcal{F}_{i}), \\
			&Pred = Soft_3(Conv2D(Cat(Up(\mathcal{D}_4[\textcolor{black}{1},:,:,:]) \\ 
			& \hspace{2.5cm}Assemble_{\forall n1, n2}(\mathcal{D}_4[\textcolor{black}{2:},:,:,:])))),
		\end{aligned}
	\end{equation}
	in which $CBR$ represents a stack of $3\times3$ Convolution, Batch Normalization, and ReLU, capable of compressing  channel dimension of tensor to $\rm{C}^{\rm{cmp}}$; $Assemble_{\forall n1, n2}$ splices the final close-view features into a full-resolution one; $Conv2D$  adjusts channel dimension to binary maps, namely foreground and background; $Soft_3$ performs exponential normalization over channel dimension to generate the final prediction.
	
	\section{Experiments}
	\label{studies}
	\subsection{RRSIS Datasets and Evaluation Metrics.}
	\textbf{Datasets:} In this work, we evaluate the performance of our method based on three challenging datasets, namely RSIS-D~\cite{liu2024rotated}, RefSegRS~\cite{yuan2024rrsis}, and RIS-Bench~\cite{crobim}. The RRSIS-D contains in total 17402 image-mask-language triplet samples, with 12181 for training, 1740 for validation, and 1817 for testing. Images have a resolution of $800\times 800$, and the pixel semantics can be categorized into 20 classes that are described by 7 attributes.  The RefSegRS 
	contains 4420 image-mask-language triplet samples, of which 2172, 413, and 1817 are used for training, validation, and testing, respectively. Image resolution in RefSegRS is $512\times 512$. Five different attribute tags and two spatial relationships are used to describe targets of 14 classes in the bundled expression. The RIS-Bench contains 52472 image-mask-language triplet samples, with 26300, 10013, and 16158 of them being used for training, validation, and testing.  Image resolution also is $512\times 512$, the pixel can be categorized into 26 classes, and target are described by 8 attributes. Statistically, the average length of language in this dataset is 14.32 words.
	
	\textbf{Metrics:} We use three widely-accepted metrics to evaluate the performance, namely overall Intersection-Over-Union (oIoU), mean Intersection-Over-Union (mIoU), and Precision@$\rm X$ ($\rm P$@$\rm X$, with the thresholding value $\rm X$ selected from the set of $\{0.5, 0.6, 0.7, 0.8, 0.9\}$). The oIoU can be obtained by calculating the ratio of total intersection region over total union region for the ground truth and corresponding prediction across all test samples. The mIoU can be computed by averaging the IoU values over the test set. The oIoU metric is sensitive to large targets due to their dominant area contribution, whereas mIoU provides equal weighting to all targets regardless of size. The $\rm Prec$@$\rm X$ reports the percentage of targets with IoU score exceeding the specified threshold value $\rm X$, reflecting the model's localization reliability.
	
	\begin{table*}[htbp]
		\caption{Quantitative comparison on RRSIS-D, in which R-101, Swin-B, and Swin-S represent ResNet-101 and Swin-Base, and Swin-Small, respectively. The top, second, and third ranked scores are marked using red, blue, and green colors, respectively.}
		\label{tab:rrsis-d}
		\fontsize{8}{12}\selectfont
		\centering
		\resizebox{\textwidth}{!}{%
			\begin{tabular}{l|c|c|c|c|c|c|c|c|c|c|c|c|c|c|c|c}
				\toprule
				\multirow{2}{*}{Method} & \makecell{Visual} & \makecell{Text} & \multicolumn{2}{c|}{Pr@0.5} & \multicolumn{2}{c|}{Pr@0.6} & \multicolumn{2}{c|}{Pr@0.7} & \multicolumn{2}{c|}{Pr@0.8} & \multicolumn{2}{c|}{Pr@0.9} & \multicolumn{2}{c|}{oIoU} & \multicolumn{2}{c}{mIoU} \\
				\cline{4-17} 
				& \makecell{Encoder} & \makecell{Encoder} & Val & Test & Val & Test & Val & Test & Val & Test & Val & Test & Val & Test & Val & Test \\
				\midrule
				LSCM \cite{hui2020linguistic} & R-101 & LSTM & 57.12 & 56.02 & 48.04 & 46.25 & 37.87 & 37.70 & 26.37 & 25.28 & 7.93 & 8.27 & 69.28 & 69.05 & 50.36 & 49.92 \\
				
				CMPC \cite{huang2020referring} & R-101 & LSTM & 57.93 & 55.83 & 48.85 & 47.40 & 38.50 & 36.94 & 25.28 & 25.45 & 9.31 & 9.19 & 70.15 & 69.22 & 50.41 & 49.24 \\
				
				BRINet \cite{hu2020bi123} & R-101 & LSTM & 58.79 & 56.90 & 49.54 & 48.77 & 39.65 & 39.12 & 28.21 & 27.03 & 9.19 & 8.73 & 70.73 & 69.88 & 51.14 & 49.65 \\
				
				CMPC+ \cite{cmpcplus} & R-101 & LSTM & 59.19 & 57.65 & 49.36 & 47.51 & 38.67 & 36.97 & 25.91 & 24.33 & 8.16 & 7.78 & 70.14 & 68.64 & 51.41 & 50.24 \\
				
				CRIS \cite{wang2022cris} & R-101 & CLIP & 56.44 & 54.84 & 47.87 & 46.77 & 39.77 & 38.06 & 29.31 & 28.15 & 11.84 & 11.52 & 70.08 & 70.46 & 50.75 & 49.69 \\
				
				LAVT \cite{yang2022lavt} & Swin-B & BERT & 69.54 & 69.52 & 63.51 & 63.61 & 53.16 & 53.29 & 43.97 & 41.60 & 24.25 & 24.94 & 77.59 & 77.19 & 61.46 & 61.04 \\
				
				CARIS \cite{liu2023caris} & Swin-B & BERT & 71.61 & 71.50 & 64.66 & 63.52 & 54.14 & 52.92 & 42.76 & 40.94 & 23.79 & 23.90 & 77.48 & 77.17 & 62.88 & 62.17 \\
				
				RIS-DMMI \cite{Hu_2023_ICCV} & Swin-B & BERT & 70.40 & 68.74 & 63.05 & 60.96 & 54.14 & 50.33 & 41.95 & 38.38 & 23.85 & 21.63 & 77.01 & 76.20 & 60.72 & 60.12 \\
				
				LGCE \cite{yuan2024rrsis} & Swin-B & BERT & 68.10 & 67.65 & 60.52 & 61.53 & 52.24 & 51.45 & 42.24 & 39.62 & 23.85 & 23.33 & 76.68 & 76.34 & 60.16 & 59.37 \\
				
				CGFormer \cite{tang2023contrastive} & Swin-B & BERT & 73.85 & 73.46 & 66.84 & 65.74 & 54.77 & 54.02 & 42.83 & 40.09 & 22.18 & 21.65 & 76.94 & 76.47 & 64.07 & 63.14 \\
				
				RefSeg \cite{wu2022towards} & Swin-B & BERT & 64.22 & 66.59 & 58.72 & 59.58 & 50.00 & 49.93 & 35.78 & 33.78 & 24.31 & 23.30 & 76.39 & 77.40 & 58.92 & 58.99 \\
				
				RMSIN \cite{liu2024rotated} & Swin-B & BERT & \textcolor{green}{\textbf{74.66}} & 74.26 & \textcolor{green}{\textbf{68.22}} & \textcolor{green}{\textbf{67.25}} & \textcolor{green}{\textbf{57.41}} & \textcolor{green}{\textbf{55.93}} & \textcolor{green}{\textbf{45.29}} & \textcolor{green}{\textbf{42.55}} & \textcolor{green}{\textbf{24.43}} & \textcolor{green}{\textbf{24.53}} & \textcolor{green}{\textbf{78.27}} & \textcolor{green}{\textbf{77.79}} & \textcolor{green}{\textbf{65.10}} & 64.20 \\
				CroBIM~\cite{crobim} & Swin-B & BERT & 74.20 & \textcolor{green}{\textbf{75.00}} & 66.15 &66.32 & 54.08 & 54.31 & 41.38 & 41.09 & 22.30 & 21.78 & 76.24 & 76.37 & 63.99 & \textcolor{green}{\textbf{64.24}}\\
				
				\hline
				CSINet & Swin-S & BERT & \textcolor{blue}{\textbf{78.51}} & \textcolor{blue}{\textbf{77.00}} & \textcolor{blue}{\textbf{71.32}} & \textcolor{blue}{\textbf{70.30}} & \textcolor{blue}{\textbf{58.45}} & \textcolor{blue}{\textbf{59.10}} & \textcolor{blue}{\textbf{45.40}} & \textcolor{blue}{\textbf{44.31}} & \textcolor{blue}{\textbf{26.21}} & \textcolor{blue}{\textbf{26.02}} & \textcolor{blue}{\textbf{78.72}} & \textcolor{blue}{\textbf{78.15}} & \textcolor{blue}{\textbf{67.59}} & \textcolor{blue}{\textbf{66.64}} \\
				CSINet & Swin-B & BERT & \textcolor{red}{\textbf{79.20}} & \textcolor{red}{\textbf{78.81}} & \textcolor{red}{\textbf{71.78}} & \textcolor{red}{\textbf{72.60}} & \textcolor{red}{\textbf{59.94}} & \textcolor{red}{\textbf{60.65}} & \textcolor{red}{\textbf{45.80}} & \textcolor{red}{\textbf{45.72}} & \textcolor{red}{\textbf{27.01}} & \textcolor{red}{\textbf{26.97}} & \textcolor{red}{\textbf{79.04}} & \textcolor{red}{\textbf{78.34}} & \textcolor{red}{\textbf{67.70}} & \textcolor{red}{\textbf{67.47}} \\
				\bottomrule
			\end{tabular}%
			\vspace{-6mm}
		}
	\end{table*}
	
	\begin{table*}[htbp]
		\caption{Quantitative comparison on RefSegRS, where R-101, Swin-B, and Swin-S represent ResNet-101 and Swin-Base, and Swin-Small, respectively. The top, second, and third ranked scores are marked as red, blue, and green, respectively.}
		\label{tab:refsegrs}
		\fontsize{8}{12}\selectfont
		\centering
		\resizebox{\textwidth}{!}{%
			\begin{tabular}{l|c|c|c|c|c|c|c|c|c|c|c|c|c|c|c|c}
				\toprule
				\multirow{2}{*}{Method} & \makecell{Visual} & \makecell{Text} & \multicolumn{2}{c|}{P@0.5} & \multicolumn{2}{c|}{P@0.6} & \multicolumn{2}{c|}{P@0.7} & \multicolumn{2}{c|}{P@0.8} & \multicolumn{2}{c|}{P@0.9} & \multicolumn{2}{c|}{oIoU} & \multicolumn{2}{c}{mIoU} \\
				\cline{4-17} 
				& \makecell{Encoder} & \makecell{Encoder} & Val & Test & Val & Test & Val & Test & Val & Test & Val & Test & Val & Test & Val & Test \\
				\midrule
				CMSA \cite{yelinwei2019cross} & R-101 & None & 39.24 & 28.07& 38.44 & 20.25& 20.39 & 12.71& 11.79 & 5.61& 1.52 & 0.83& 65.84& 64.53& 43.62 & 41.47\\
				RRN \cite{li2018referring} & R-101 & LSTM & 55.43 & 30.26 & 42.98 & 23.01 & 23.11 & 14.87 & 13.72 & 7.17 & 2.64 & 0.98 & 69.24 & 65.06 & 50.81 & 41.88 \\
				CMPC+ \cite{cmpcplus} & R-101 & LSTM &56.84&  49.19&  37.59&  28.31&  20.42&  15.31&  10.67&  8.12&  2.78&  2.55&  70.62&  66.53&  47.13&  43.65  \\
				CMPC \cite{huang2020referring} & R-101 & LSTM &  46.09&  32.36&  26.45&  14.14&  12.76&  6.55&  7.42&  1.76&  1.39&  0.22&  63.55&  55.39&  42.08&  40.63\\
				CRIS \cite{wang2022cris} & R-101 & CLIP & 53.13 & 35.77 & 36.19 & 24.11 & 24.36 & 14.36 & 11.83 & 6.38 & 2.55 & 1.21 & 72.14 & 65.87 & 53.74 & 43.26 \\
				
				LAVT \cite{yang2022lavt} & Swin-B & BERT & 80.97 & 51.84 & 58.70 & 30.27 & 31.09 & 17.34 & 15.55 & 9.52 & 4.64 & 2.09 & 78.50 & 71.86 & 61.53 & 47.40 \\
				CARIS \cite{liu2023caris} & Swin-B & BERT & 68.45 & 45.40 & 47.10 & 27.19 & 25.52 & 15.08 & 14.62 & 8.87 & 3.71 & 1.98 & 75.79 & 69.74 & 54.30 & 42.66 \\
				RIS-DMMI \cite{Hu_2023_ICCV} & Swin-B & BERT & 86.17 & 63.89 & 74.71 & 44.30 & 38.05 & 19.81 & 18.10 & 6.49 & 3.25 & 1.00 & 74.02 & 68.58 & 65.72 & 52.15 \\
				LGCE \cite{yuan2024rrsis} & Swin-B & BERT &  90.72&  73.75&  86.31&  61.14&  71.93&  39.46&  32.95&  16.02&  \textcolor{green}{\textbf{10.21}}&  \textcolor{green}{\textbf{5.45}} &  \textcolor{green}{\textbf{83.56}} &  \textcolor{green}{\textbf{76.81}} &  72.51&  59.96\\
				CGFormer \cite{tang2023contrastive} & Swin-B & BERT &  \textcolor{green}{\textbf{94.68}} &  78.11&  86.81& \textcolor{green}{\textbf{67.27}}&  71.30&  40.54&  24.77&  16.17&  4.63& 1.93&  78.08&  73.19&  72.19& 61.79\\
				
				RefSeg \cite{wu2022towards} & Swin-B & BERT & 81.67 & 50.25 & 52.44 & 28.01 & 30.86 & 17.83 & 17.17 & 9.19 & 5.80 & 2.48 & 77.74 & 71.13 & 60.44 & 47.12 \\
				
				RMSIN \cite{liu2024rotated} & Swin-B & BERT &  93.97&  \textcolor{green}{\textbf{79.20}} &  \textcolor{green}{\textbf{89.33}} &  65.99&  \textcolor{green}{\textbf{74.25}}&  \textcolor{green}{\textbf{42.98}} &  \textcolor{green}{\textbf{29.70}} &  \textcolor{green}{\textbf{16.51}} &  7.89&  3.25&  82.41&  75.72&  \textcolor{green}{\textbf{73.84}} &  \textcolor{green}{\textbf{62.58}} \\
				CroBIM~\cite{crobim} & Swin-B & BERT & 87.24 & 64.83 & 75.17 & 44.41 & 44.78 & 17.28 & 19.03 & 9.69 & 6.26 & 2.20&  78.85 & 72.30 & 65.79 & 52.69 \\	
				\hline
				CSINet & Swin-S & BERT & \textcolor{blue}{\textbf{96.51}} & \textcolor{blue}{\textbf{86.46}} & \textcolor{blue}{\textbf{95.60}} & \textcolor{blue}{\textbf{80.79}} & \textcolor{blue}{\textbf{92.13}} & \textcolor{blue}{\textbf{69.90}} & \textcolor{blue}{\textbf{76.62}} & \textcolor{blue}{\textbf{41.17}} & \textcolor{blue}{\textbf{16.67}} & \textcolor{blue}{\textbf{8.86}} & \textcolor{blue}{\textbf{87.25}} & \textcolor{blue}{\textbf{81.32}} & \textcolor{blue}{\textbf{81.92}} & \textcolor{blue}{\textbf{71.14}} \\
				CSINet & Swin-B & BERT & \textcolor{red}{\textbf{96.53}} & \textcolor{red}{\textbf{87.45}} & \textcolor{red}{\textbf{95.83}} & \textcolor{red}{\textbf{83.21}} & \textcolor{red}{\textbf{94.91}} & \textcolor{red}{\textbf{74.96}} & \textcolor{red}{\textbf{89.35}} & \textcolor{red}{\textbf{54.60}} & \textcolor{red}{\textbf{25.00}} & \textcolor{red}{\textbf{13.70}} & \textcolor{red}{\textbf{89.14}} & \textcolor{red}{\textbf{83.43}} & \textcolor{red}{\textbf{84.32}} & \textcolor{red}{\textbf{73.88}} \\
				\bottomrule
			\end{tabular}%
			\vspace{-6mm}
		}
	\end{table*}

	\begin{table*}[htbp]
		\centering
		\caption{Quantitative comparison on RIS-Bench, where R-101, Swin-B, and Swin-S represent ResNet-101 and Swin-Base, and Swin-Small, respectively. The top, second, and third ranked scores are marked as red, blue, and green, respectively.}
		\label{tab:risbench}
		\fontsize{8}{12}\selectfont
		\resizebox{\textwidth}{!}{
			\begin{tabular}{l|c|c|c|c|c|c|c|c|c|c|c|c|c|c|c|c}
				\cmidrule{1-17} 
				\multirow{2}{*}{Method} & \makecell{Visual} & \makecell{Text} & \multicolumn{2}{c|}{Pr@0.5} & \multicolumn{2}{c|}{Pr@0.6} & \multicolumn{2}{c|}{Pr@0.7} & \multicolumn{2}{c|}{Pr@0.8} & \multicolumn{2}{c|}{Pr@0.9} & \multicolumn{2}{c|}{oIoU} & \multicolumn{2}{c}{mIoU} \\ \cline{4-17}  
				& \makecell{Encoder} & \makecell{Encoder} & Val & Test & Val & Test & Val & Test & Val & Test & Val & Test & Val & Test & Val & Test \\ 
				\cmidrule{1-17} 
				MAttNet\cite{yu2018mattnet} & R-101 & LSTM & 56.77 & 56.83 & 48.51 & 48.02 & 41.53 & 41.75 & 34.33 & 34.18 & 13.84 & 15.26 & 48.66 & 51.24 & 44.28 & 45.71 \\
				CMPC\cite{huang2020referring} & R-101 & LSTM & 54.89 & 55.17 & 47.77 & 47.84 & 40.38 & 40.28 & 32.89 & 32.87 & 12.63 & 14.55 & 47.59 & 50.24 & 42.83 & 43.82\\
				CMPC+\cite{cmpcplus} & R-101 & LSTM & 57.84 & 58.02 & 49.24 & 49.00 & 42.34 & 42.53 & 35.77 & 35.26 & 14.55 & 17.88 & 50.29 & 53.98 & 45.81 & 46.73 \\	
				ETRIS\cite{xu2023bridging} & R-101 & CLIP & 59.87 & 60.98 & 49.91 & 51.88 & 35.88 & 39.87 & 20.10 & 24.49 & 8.54 & 11.18 & 64.09 & 67.61 & 51.13 & 53.06 \\
				CRIS\cite{wang2022cris} & R-101 & CLIP & 63.42 & 63.67 & 54.32 & 55.73 & 41.15 & 44.42 & 24.66 & 28.80 & 10.27 & 13.27 & 66.26 & 69.11 & 53.64 & 55.18 \\
				LAVT\cite{yang2022lavt} & Swin-B & BERT & 68.27 & 69.40 & 62.71 & 63.66 & 54.46 & 56.10 & 43.13 & 44.95 & 21.61 & 25.21 & 69.39 & 74.15 & 60.45 & 61.93 \\
				
				CrossVLT\cite{vlt} & Swin-B & BERT & 70.05 & 70.62 & 64.29 & 65.05 & 56.97 & 57.40 & 44.49 & 45.80 & 21.47 & 26.10 & 69.77 & 74.33 & 61.54 & 62.84 \\ 
				RIS-DMMI\cite{Hu_2023_ICCV} & Swin-B & BERT & 71.27 & 72.05 & 66.02 & 66.48 & 58.22 & 59.07 & 45.57 & 47.16 & 22.43 & 26.57 & \textcolor{blue}{\textbf{70.58}} & \textcolor{blue}{\textbf{74.82}} & 62.62 & 63.93 \\
				
				LGCE\cite{yuan2024rrsis} & Swin-B & BERT & 68.20 & 69.64 & 62.91 & 64.07 & 55.01 & 56.26 & 43.38 & 44.92 & 21.58 & 25.74 & 68.81 & 73.87 & 60.44 & 62.13 \\ 
				
				RefSeg~\cite{wu2022towards} & Swin-B & BERT & 67.42 & 69.15 & 61.72 & 63.24 & 53.64 & 55.33 & 40.71 & 43.27 & 19.43 & 24.20 & 69.50 & 74.23 & 59.37 & 61.25 \\

				CGFormer\cite{tang2023contrastive} & Swin-B & BERT & 74.92 & 73.38 & 69.41 &66.42 & 61.98 &57.75 & 48.35 & 44.15 & 21.02 & 21.81 & 68.68 & 73.11 & 65.41 & 64.95 \\
				RMSIN\cite{liu2024rotated} & Swin-B & BERT & 70.05 & 71.01 & 64.64 & 65.46 & 56.37 & 57.69 & 44.14 & 45.50 & 21.40 & 25.92 & 69.51 & 74.09 & 61.78 & 63.07 \\
				
				CroBIM~\cite{crobim}  & Swin-B & BERT & \textcolor{blue}{\textbf{76.59}} & \textcolor{green}{\textbf{75.75}} & \textcolor{green}{\textbf{71.73}} & \textcolor{green}{\textbf{70.34}} & \textcolor{green}{\textbf{64.32}} & \textcolor{green}{\textbf{63.12}} & \textcolor{green}{\textbf{53.18}} & \textcolor{green}{\textbf{51.12}} & \textcolor{green}{\textbf{28.53}} & \textcolor{green}{\textbf{28.45}} & 69.08 & 73.61 & \textcolor{green}{\textbf{67.52}} & \textcolor{green}{\textbf{67.32}} \\	
				
				\hline
				CSINet (Ours) & Swin-S & BERT & \textcolor{green}{\textbf{76.37}} & \textcolor{blue}{\textbf{76.45}} & \textcolor{blue}{\textbf{71.76}} & \textcolor{blue}{\textbf{71.58}} & \textcolor{blue}{\textbf{65.15}} & \textcolor{blue}{\textbf{64.94}} & \textcolor{blue}{\textbf{54.77}} & \textcolor{blue}{\textbf{54.41}} & \textcolor{blue}{\textbf{31.98}} & \textcolor{blue}{\textbf{34.09}} & \textcolor{green}{\textbf{70.36}} & \textcolor{green}{\textbf{74.74}} & \textcolor{blue}{\textbf{67.92}} & \textcolor{blue}{\textbf{68.43}}  \\	
				CSINet (Ours) & Swin-B & BERT & \textcolor{red}{\textbf{77.17}} & \textcolor{red}{\textbf{77.12}} & \textcolor{red}{\textbf{72.50}} & \textcolor{red}{\textbf{72.40}} & \textcolor{red}{\textbf{66.44}} & \textcolor{red}{\textbf{66.04}} & \textcolor{red}{\textbf{56.17}} & \textcolor{red}{\textbf{55.86}} & \textcolor{red}{\textbf{33.01}} & \textcolor{red}{\textbf{35.34}} & \textcolor{red}{\textbf{71.16}} & \textcolor{red}{\textbf{75.36}} & \textcolor{red}{\textbf{68.86}} & \textcolor{red}{\textbf{69.25}}  \\	
				\hline
			\end{tabular}
		}
	\end{table*}
	\begin{table}[htbp]
		\centering
		\vspace{-2mm}
		\caption{MIoU on each category of RRSIS-D Dataset. }
		\label{tab:categories}
		\resizebox{0.49\textwidth}{!}{
			\setlength\tabcolsep{3pt}
			\renewcommand\arraystretch{1}
			\begin{tabular}{@{}l|c|c|c|c|c@{}}
				\toprule
				\textbf{Category}                & LAVT & LGCE & CGFormer & RMSIN & CSINet \\ \midrule
				Airport                         & 62.71     &64.19 &67.92   & 66.13& 70.43\\
				Golf field                      & 71.72  & 71.79   &72.96     & 73.30& 74.73\\
				Expressway service area         & 65.41  & 66.87    &69.66    & 68.69& 69.80\\
				Baseball field                  & 75.92 &74.38 &79.27 & 79.13& 82.06\\ 
				Stadium                         & 76.20 & 75.93 &78.30 & 78.79& 83.04\\ 
				Ground track field              & 69.39  & 67.95   &74.08     & 73.47& 77.95\\ 
				Storage tank                    & 71.33 & 71.16   &75.97       & 76.54& 77.65\\ 
				Basketball court                & 68.36 & 68.95 & 70.20& 70.06& 74.74\\ 
				Chimney                         & 68.64 & 68.25     &71.27     & 69.79& 74.78\\ 
				Tennis court                    & 64.21  & 63.92     &70.72   & 68.31& 74.78\\ 
				Overpass                        & 54.74  & 55.62    &57.45     & 59.34& 62.09\\ 
				Train station                   & 58.40 & 58.18  &60.56& 58.15& 61.01\\ 
				Ship                            & 59.41 & 60.52 &62.70 & 63.17& 69.19\\ 
				Expressway toll station         & 53.75 & 54.85  &60.73 & 61.45& 69.34\\ 
				Dam                             &57.55     & 58.44   &60.36   & 60.21& 64.31\\ 
				Harbor                          & 35.03    & 34.09  &37.74     & 37.63& 46.24\\
				Bridge                          & 42.30      & 44.51  &48.63   & 49.64& 54.92\\
				Vehicle                         & 37.63  & 41.36     &48.97   & 49.22& 55.28\\ 
				Windmill                        & 57.32    & 57.04   &59.30    & 59.66& 61.66\\ \midrule
				\textbf{Average}                & 60.53          &60.95 & 64.57 & 64.35 & 68.63\\ \bottomrule
		\end{tabular}}
	\end{table}
	
	\subsection{Implementation Details.}
	We conduct all experiments using Pytorch and HuggingFace frameworks on a workstation with two GTX 3090 GPUs. The pre-trained weights of Swin Transformer are generated in the classification task of ImageNet22K. Other parameters are initialized randomly. The 12-layer Bert with hidden dimension of 768 is initialized with the HuggingFace-provided weights. The clipping length of the language is set to 20. We use AdamW optimizer with weight decay 0.01, an initial learning rate 0.00005, and polynomial learning rate decay strategy to drive the parameter updating process. We use Dice loss and BCE loss to supervise the output of the model, and set their weights to 0.9 and 0.1 in loss calculation, respectively. For RRSIS-D, RefSegRS, and RIS-Bench, the training epochs are set to 40, 60, and 40, respectively. Batch size is set to 8 for all experiments. Both the image in remote view and parch images in close view are resized to $384\times 384$ before fed into the network. In CVWin, the hyper-parameter $\rm{S}^{win}$ will keep aligned to that in Swin Transformer~\cite{swin}. Inspired by ASPP~\cite{aspp}, the density number $\rm{J}$ of collaboratively dilation attention is set to 3. Empirically, the slice size in Collaborative Dilated Attention is set to $5$. The super-parameter $\rm{C}^{cmp}$ in CDAD decoder is set to $\rm{C}_4^{\rm{vis}}/2$ in Eq. \eqref{remaining_decoder}.

	\subsection{Comparison with the SOTAs.}
	We document the results of P@X, oIoU and mIoU scores on RRSIS-D, RefSegRS, and RIS-Bench in Table \ref{tab:rrsis-d}, Table \ref{tab:refsegrs}, and Table \ref{tab:risbench}, respectively. 
    The results of CMSA \cite{yelinwei2019cross},
	RRN \cite{li2018referring}, CMPC+ \cite{cmpcplus}, CMPC \cite{huang2020referring},
	CRIS \cite{wang2022cris}, LAVT \cite{yang2022lavt}, CARIS \cite{liu2023caris}, RIS-DMMI \cite{Hu_2023_ICCV},
	LGCE \cite{yuan2024rrsis},
	CGFormer \cite{tang2023contrastive},
	RefSeg \cite{wu2022towards},
	RMSIN \cite{liu2024rotated},
	and CroBIM~\cite{crobim} are also included. 
	In the tables, R-101, Swin-S, and Swin-B represent ResNet-101, Swin-Small Transformer, and Swin-Base Transformer backbones. Bert base is used to encode the language. It can be seen that our network can consistently generate satisfactory performance across different datasets.
	\begin{figure*}[htbp]
		\centering
		\includegraphics[scale=0.89]{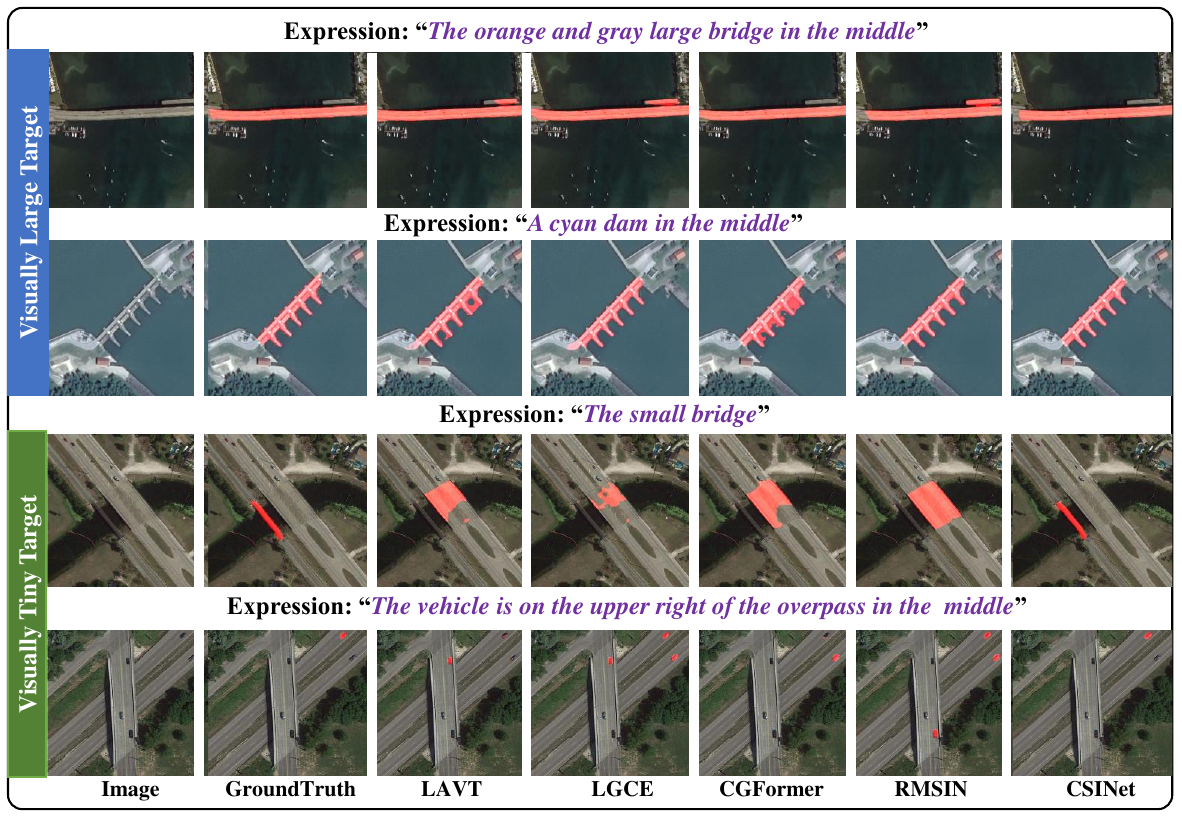}
		\caption{Visualization results of proposed CSINet and other state of the art algorithms from the validation set of RRSIS-D.}
		\label{Comparison}
	\end{figure*}
	
	\begin{table*}[htbp]
		\centering
		\caption{Ablation studies on the validation set of RRSIS-D. The details of CVWin and CDAD are investigated.}
		\resizebox{0.84\textwidth}{!}{
			\setlength\tabcolsep{3pt}
			\renewcommand\arraystretch{1}
			\begin{tabular}{c| c| c c c c c |c c}
				\toprule
				Experiment &Method &P@0.5$\uparrow$  &P@0.6$\uparrow$ &P@0.7$\uparrow$ &P@0.8$\uparrow$ &P@0.9$\uparrow$ &oIoU$\uparrow$ &mIoU$\uparrow$ \\
				\midrule
				$\#1$ &CVWin+CDAD (Full-packaged) &78.51 &71.32 &58.45 &45.40 &26.21 &78.72 &67.59\\
				\midrule
				$\#1.1$& PWAM~\cite{yang2022lavt} +CDAD&76.03 &68.45 &56.38 & 43.68 &25.06 &77.66 &65.99\\
				$\#1.2$& IIM~\cite{liu2024rotated}+CDAD &76.78 &69.37 &57.30 & 44.77 &25.75 &77.42 &66.52\\
				$\#1.3$& Direct Sum+CDAD &75.86 &68.85 &57.87 &44.08 &25.98 &77.83 &65.75\\
				$\#1.4$& CVWin (-Gate)+CDAD &77.18 &69.89 &58.45 &44.14 &26.32 &78.58 &66.93\\
				\midrule
				$\#2.1$& CVWin+ARC~\cite{liu2024rotated} &74.37 &66.72 &55.52 &44.05 &24.31 &77.38 &65.02\\
				$\#2.2$& CVWin+CDAD (-CDA-Skip) &76.03 & 68.39 &57.13 &43.62 &24.60 &77.57 &66.07\\
				$\#2.3$& CVWin+CDAD (-Skip) &77.01 &69.08 &56.67 &44.83 &25.23 &77.36 &66.55\\
				\bottomrule
		\end{tabular}}	
		\label{tab:ablation}
	\end{table*}
	
	Although recent transformer-based methods have demonstrated substantial advancements over conventional CNN-based algorithms, our proposed CSINet exhibits even more pronounced performance gains.
    As evidenced by the results on the RRSIS-D dataset in Table \ref{tab:rrsis-d}, the Swin-Base-powered CSINet outperforms the previous CNN-based state-of-the-art CRIS by 20.40, 7.76, and 14.43 points in average P@X, oIoU, and mIoU metrics, respectively.  On the same feature extractor Swin-Base Transformer, our CSINet surpasses current RIS-leading method CARIS by 5.87, 1.37, and 5.05 points against average P@X, oIoU, and mIoU. Besides, in comparison to RMSIN that is previously tailored for RRSIS task, our CSINet still can deliver consistent enhancements of 3.40, 0.48, and 2.94 points on P@X, oIoU, and mIoU under identical backbone configurations. Notably, the improvement in mIoU by our CSINet over other state-of-the-art methods are more pronounced than that in oIoU, highlighting our method's strong capability in handling tiny targets.

	As demonstrated in Table \ref{tab:refsegrs} and Table \ref{tab:risbench}, our method exhibits robust cross-dataset generalization capabilities. The training part in RefSegRS is only 1/6 size of RRSIS-D in terms of training sample number. Specifically, on the RefSegRS benchmark with merely 1/6 of RRSIS-D's training samples, CSINet achieves state-of-the-art performance. Notably although LGCE optimizes its techniques specifically on RefSegRS dataset, our CSINet with Swin-Small Transformer can surpass it by 17.68, 4.09, and 10.3 points in average P@X, oIoU, and mIoU, respectively. Compared to RMSIN that  which utilizes the larger Swin-Base visual encoder, our CSINet with Swin-Small Transformer surpasses the method by Under identical backbone conditions (Swin-Base), CSINet's superiority over RMSIN becomes more pronounced, delivering 21.25, 7.22, and 10.89 gains across the three metrics. The RIS-Bench dataset has two times more training samples than that of RRSIS-D, and more intricate language expression. Our CSINet still can persist superior performance over others. Our Swin-Small variant outperforms Swin-Base-based CroBIM by 1.84, 1.21, and 0.76, while the same-backbone comparison reveals 2.89, 1.95, and 1.64 improvements.
	
	To further benchmark the proposed method, mIoU for target in each category is documented in Table \ref{tab:categories}, which includes samples in both validation and test sets of the RRSIS-D dataset. All methods use Swin-Base Transformer and BERT-Base to extract vision and language semantics. From the table, it can be seen that our method CSINet can achieve satisfactory performance across diverse categories, especially for small targets. In the Chimney, Ship, and Vehicle categories, CSINet achieves improvements of 4.99, 6.02, and 6.06, respectively, highlighting its precision in localizing small targets. Figure \ref{Comparison} visualizes targets with varying geometric scales, where all compared methods use the Swin-Base backbone. The first two rows display large-scale targets, with CSINet producing more continuous and accurate segmentation boundaries. The last two rows focus on tiny targets, where CSINet effectively distinguishes targets from background regions with similar textures, demonstrating superior robustness to visual ambiguities.
	\begin{table}[htbp]
		\centering
		\caption{The impact of close-view partition pattern.}
		\resizebox{0.48\textwidth}{!}{
			\setlength\tabcolsep{3pt}
			\renewcommand\arraystretch{1}
			\begin{tabular}{c|ccccc}
				\toprule
				($\rm{N}^{\rm{view}},\rm{H},\rm{W}$) &P@0.5$\uparrow$  &P@0.7$\uparrow$ &P@0.9$\uparrow$ &oIoU$\uparrow$ &mIoU$\uparrow$ \\
				\midrule
				$(2, 384, 384)$ &78.51 &58.45 &26.21 &78.72 &67.59\\
				$(3,256,256)$ &74.37 &54.66 &24.37 &77.57 &64.85\\
				$(4,192,192)$ &72.30 &52.76 &22.59 &76.64 &63.23\\
				\bottomrule
		\end{tabular}}	
		\label{tab:view}
	\end{table}

	\begin{table}[htbp]
		\centering
		\caption{Ablation studies on the validation set of RRSIS-D. The details of cross-view interaction are investigated.}
		\resizebox{0.50\textwidth}{!}{
			\setlength\tabcolsep{3pt}
			\renewcommand\arraystretch{1}
			\begin{tabular}{c|ccccc}
				\toprule
				Interaction Strategy &P@0.5$\uparrow$  &P@0.7$\uparrow$ &P@0.9$\uparrow$ &oIoU$\uparrow$ &mIoU$\uparrow$ \\
				\midrule
				Close+Remote &78.51 &58.45 &26.21 &78.72 &67.59\\
				Only Close &59.94 &40.11 &16.78 &60.67 &54.63\\
				Only Remote &73.33 &52.99 &22.41 &77.47 &63.54\\
				Only Remote2Close &75.17 &55.86  &24.71 &77.93 &64.95\\
				Only Close2Remote &74.37  &55.75 &23.85 &77.56 &65.14 \\
				\bottomrule
		\end{tabular}}	
		\label{tab:interaction}
	\end{table}
	\subsection{Ablation Studies.}
	Following RMSIN~\cite{liu2024rotated}, we validate the effectiveness of our proposed techniques on the validation part of RRSIS-D dataset. To speed up the whole process, we employ Swin-Small Transformer to extract the vision semantics in the following involved experiments if no further specifications. 
	
	\textbf{Effectiveness of Architecture.} 
    To evaluate the impact of close-view partitioning strategies, we compare the default $2\times 2$ split pattern with finer-grained divisions, namely $3\times 3$ and $4\times 4$ patches, in the second and third rows of Table~\ref{tab:view}. The results demonstrate that current $2\times 2$ configuration achieves optimal performance over others.
	\begin{table}[tbp]
		\centering
		\caption{Impact of decoder stage on RRSIS-D val.}
		\resizebox{0.45\textwidth}{!}{
			\setlength\tabcolsep{3pt}
			\renewcommand\arraystretch{1}
			\begin{tabular}{c|ccccc}
				\toprule
				Method &P@0.5$\uparrow$  &P@0.7$\uparrow$ &P@0.9$\uparrow$ &oIoU$\uparrow$ &mIoU$\uparrow$ \\
				\midrule
				Full &78.51 &58.45 &26.21 &78.72 &67.59\\
				-D4 &76.21 &55.57 &23.73 &77.69 &67.63\\
				-(D4,D3) &73.45 &54.31 &22.36 &77.05 &64.20\\
				-(D4,D3,D2) &64.37 &46.26 &17.07 &75.99 &59.20\\
				\bottomrule
		\end{tabular}}	
		\label{tab:stage}
	\end{table}
	\begin{table}[tbp]
		\centering
		\caption{Impact of slice size in CDA on RRSIS-D val.}
		\resizebox{0.44\textwidth}{!}{
			\setlength\tabcolsep{3pt}
			\renewcommand\arraystretch{1}
			\begin{tabular}{lcccc}
				\toprule
				Slice Size &J=3  &J=4 &J=5 &J=6\\
				\midrule
				Overall IoU (CSINet small)&77.73  &78.39 &78.72 &78.31\\
				Mean IoU (CSINet small) & 65.33 & 67.11 &67.59 &66.76\\
				\bottomrule
		\end{tabular}}	
		\label{tab:slicesize}
	\end{table}
	\begin{table}[tbp]
		\centering
		\caption{Inference speed and parameter tested on RRSIS-D.}
		\resizebox{0.48\textwidth}{!}{
			\setlength\tabcolsep{3pt}
			\renewcommand\arraystretch{1}
			\begin{tabular}{ccccc}
				\toprule
				Method &CGFormer  &RMSIN &CSINet (Small) &CSINet\\
				\midrule
				Params &217.9 (M) &208.9 (M)&194.4 (M)&278.4 (M)\\
				Speed & 23.3 (FPS)& 21.0 (FPS)&17.1 (FPS)&15.6 (FPS)\\
				\bottomrule
		\end{tabular}}	
		\label{tab:statistics}
	\end{table}
	
\begin{figure}[tbp]
		\centering
		\includegraphics[scale=0.75]{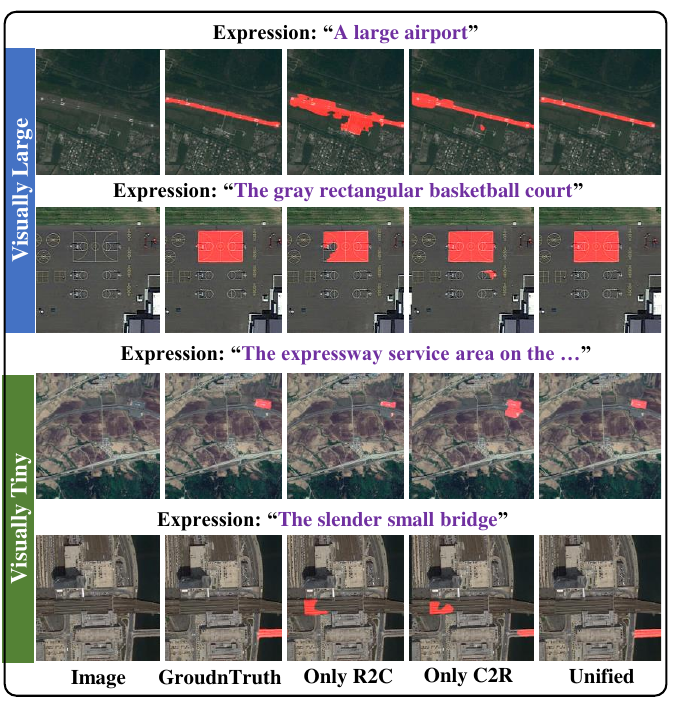}
		\caption{Quantitative results. The symbols Only R2C and Only C2R represent supplementing global information and detail cues into close-view and remote-view branches, respectively.} 
		\label{fig:unilaterally}
	\end{figure}

    Additionally, to investigate cross-view interaction strategy, we disable the information flow between the remote and close view branches. The results are documented in entries of "Only Close" and "Only Remote" in Table \ref{tab:interaction}, respectively. It can be seen that the unified framework surpasses these single-view variants by 15.45, 18.04, 12.96 (only close view) and 4.81, 1.25, 4.05 (only remote view) on average P@X, oIoU, and mIoU, respectively. We further examine unidirectional interaction solutions, namely "Only Remote-to-Close" and "Only Close-to-Remote" information flow, and document the results in Table \ref{tab:interaction}.  The solutions of "Only Remote-to-Close" and "Only Close-to-Remote" can achieve 12.97, 17.26, 10.32 and 1.75, 0.09, 1.6 better performance than the single view solutions of "Only Close" and "Only Remote", respectively. These results validate the complementary nature of bidirectional feature exchange. In Figure \ref{fig:unilaterally}, some quantitative results are shown. As demonstrated, the results by current bidirectional cross-view strategy are more delicate. 
	
	\textbf{Effectiveness of CVWin attention.} To investigate the effectiveness of our cross-view interaction strategy, we replace CVWin with the independent cross-modality fusion method PWAM from LAVT~\cite{yang2022lavt} and IIM from RMSIN~\cite{liu2024rotated}, of which the results are documented in experiment $\#1.1$ and $\#1.2$ of Table~\ref{tab:ablation}. Based on the same decoder CDAD, our method CVWin surpasses PWAM and IIM by 2.06, 1.06, 1.6 and 1.18, 1.3, 1.07 on average P@X, oIoU, mIoU, respectively. In $\#1.3$, we replace CVWin with coarse alignment, namely bilinear sampling. The CVWin achieves better performance of 1.45, 0.89, and 1.84 points over the coarse strategy. To investigate the effectiveness of stabilizing function, namely Gate function, In $\#1.4$, we replace the manipulation with direct summation in CVWin's cross-modality feature integration. It can be seen that the design will bring gains of 0.78, 0.14, and 0.66 point. These results demonstrate that our full-packaged CVWin achieves optimal performance.
	
	\textbf{Effectiveness of CDAD.} To validate the overall effectiveness of CDAD, we replace the proposed decoder with the ARC decoder adopted in RMSIN~\cite{liu2024rotated}.  The corrresponding results are documented in the entry of $\#2.1$ in Table \ref{tab:ablation}. Replacing CDAD with the ARC decoder~\cite{liu2024rotated} causes performance degradation of 2.98 on average P@X, 0.94 on oIoU, and 2.57 on mIoU, confirming CDAD's superiority in cross-view and cross-scale semantic integration. To investigate the effectiveness of Collaboratively Dilated Attention, we remove the enhancement strategy from the decoder, and document its results in $\#2.2$. As demonstrated, the removal will result in a reduction of 2.02, 1.15, and 1.52 points against the three corresponding metrics. To validate the skip-connection enhancement in CDAD decoder, we remove it and document the results in entry $\#2.3$.  On the three metrics, the design produces better performance of 1.41, 1.36 and 1.04 point gains. To investigate the contribution of multiscale feature integration, we gradually remove the decoding stage of CDAD in reverse order. The results are documented in Table \ref{tab:stage}. From the table, it can be seen that incorporation of the 1st-to-4th vision features into decoding process is vital to the capturing of both large and tiny target appearances, particularly for tiny targets.
	
	\textbf{Inference Speed and Parameters.} We also show the inference speed and parameter amount of CGFormer,  RMSIN, and our CSINet in Table \ref{tab:statistics}. Note that other than CSINet (Small) that uses Swin-Small Transformer to encode vision feature, other methods are built based on Swin-Base Transformer. The FPS (Frame Per Second) value for all of the methods is tested on a RTX 3090 GPU card with automatic mixed precision enabled. From the table, it can be seen that thanks to the strong ability of GPU's parallel computation, CSINet (Small) only lags behind RMSIN by 3.9 FPS, while requiring 14.5M less parameter and achieving significant improvements over others in accuracy across different datasets. 
	
	\section{Conclusion}
	In this work, a unified parallel segmentation framework Cross-view Semantics Interaction Network (CSINet) is proposed to perform effective referring remote sensing image segmentation. It takes dual-view inputs to extract global and local cues, and employs a cross-view window cross-attention module (CVWin) to enable bidirectional interaction across the different-propensity features. It supplements local and global information into distant and close-view features, respectively. To mine the orientational property of target, a Collaboratively Dilated Attention Decoder (CDAD) is employed while integrating cross-view multi-scale features. The CSINet unifies global and local semantically-enriched features in a seamless manner, yielding satisfactory performance. 
	
	\textbf{Limitations and Future Works.} The CSINet has the following limitations: (1) it can only handle samples that require language description closely related to the image. Integrating image-level relevance prediction and engineering more complicate dataset should be desired in the future works.  (2) current RRSIS dataset is quite constrained by the design philosophy of RIS dataset. High resolution datasets should be engineered. In the future, method should be able to strike a balance between accuracy and speed on such dataset. (3) The language encoder (\emph{e.g.,} BERT-Base) lacks the cross-modal guidance from vision branch. Enhancing its ability to capture vision-critical content requires to design effective deeper vision-language interaction mechanisms.
	
	\bibliographystyle{IEEEtran}
	\bibliography{egbib}
\end{document}